\documentclass[10pt,twocolumn,letterpaper]{article}

\usepackage{cvpr}              % To produce the CAMERA-READY version
\usepackage{graphicx}
\usepackage{amsmath}
\usepackage{amssymb}
\usepackage{booktabs}
\usepackage{stfloats}
\usepackage{amsfonts}       % blackboard math symbols
\usepackage{nicefrac}       % compact symbols for 1/2, etc.
\usepackage{microtype}      % microtypography
\usepackage{xcolor}         % colors
\usepackage{graphicx}
\usepackage{diagbox}
\usepackage{multicol}
\usepackage{enumerate}
\usepackage{times}
\usepackage{epsfig}
\usepackage{graphicx}
\usepackage{amsmath}
\usepackage{amssymb}
\usepackage{threeparttable}
\usepackage{enumitem}
\usepackage{multirow}
\usepackage{color}
\usepackage{array}

\usepackage{ulem}

\usepackage[colorlinks,linkcolor=red]{hyperref}
\usepackage[numbers,sort&compress]{natbib}
\setenumerate[1]{leftmargin = 10pt, itemsep=0pt,partopsep=0pt,parsep=-0pt,topsep=-0pt}
\usepackage{hyperref}

\newcommand{\PreserveBackslash}[1]{\let\temp=\\#1\let\\=\temp}

\usepackage[capitalize]{cleveref}
\crefname{section}{Sec.}{Secs.}
\Crefname{section}{Section}{Sections}
\Crefname{table}{Table}{Tables}
\crefname{table}{Tab.}{Tabs.}

\def\cvprPaperID{5008} % *** Enter the CVPR Paper ID here
\def\confName{CVPR}
\def\confYear{2022}
\def\hlinew#1{%
 \noalign{\ifnum0=`}\fi\hrule \@height #1 \futurelet
 \reserved@a\@xhline}
 
\begin{document}


\title{1st Solution Places for CVPR 2023 UG$^{\textbf{2}}$+ Challenge Track 2.1-\\Text Recognition through Atmospheric Turbulence}

\author{Shengqi Xu, Xueyao Xiao, Shuning Cao, Yi Chang\footnotemark[1],  Luxin Yan\\
National Key Lab of Multispectral Information Intelligent Processing Technology,\\
Huazhong University of Science and Technology, China\\
{\tt\small \{m202273123, xiaoxueyao, sn\_cao, yichang, yanluxin\}@hust.edu.cn}}
\maketitle

%%%%%%%%% ABSTRACT
\begin{abstract}
  In this technical report, we present the solution developed by our team “VIELab-HUST” for text recognition through atmospheric turbulence in Track 2.1 of the CVPR 2023 UG$^{2}$+ challenge. Our solution involves an efficient multi-stage framework that restores a high-quality image from distorted frames. Specifically, a frame selection algorithm based on sharpness is first utilized to select the sharpest set of distorted frames.
  Next, each frame in the selected frames is aligned to suppress geometric distortion through optical-flow-based image registration. Then, a region-based image fusion method with DT-CWT is utilized to mitigate the blur caused by the turbulence. Finally, a learning-based deartifacts method is applied to remove the artifacts in the fused image, generating a high-quality outuput. Our framework can handle both hot-air text dataset and turbulence text dataset provided in the final testing phase and \textbf{achieved 1st place in text recognition accuracy}. Our code will be available at \url{{https://github.com/xsqhust/Turbulence_Removal}}.

%    In this technical report, we briefly introduce the solution
% of our team HUST\li VIE for GT-Rain Challenge in CVPR 2023 UG$^{2}$+ Track 3. In this task,
% we propose an efficient two-stage framework to reconstruct
% a clear image from rainy frames. Firstly, a low-rank based video deraining method is utilized to generate pseudo GT, which fully takes the advantage of multi and aligned rainy frames. Secondly, a transformer-based single image deraining network Uformer is implemented to pre-train on large real rain dataset and then fine-tuned on pseudo GT to further improve image restoration. Moreover, in terms of visual pleasing effect, a comprehensive image processor module is utilized at the end of pipeline. Our overall framework is elaborately designed and able to handle both heavy rainy and foggy sequences provided in the final testing phase. Finally, we rank \textbf{1st on the average structural similarity (SSIM) and rank 2nd on the average peak signal-to-noise ratio (PSNR)}. Our code is available at \url{{https://github.com/yunguo224/UG2_Deraining}}.
\end{abstract}

%%%%%%%%% BODY TEXT
\vspace{-10pt}
\section{Introduction}
\begin{figure}[t]
  \vspace{0cm}  %调整图片与上文的垂直距离
 \setlength{\abovecaptionskip}{0 cm}   %调整图片标题与图距离
 \setlength{\belowcaptionskip}{-0.4 cm}   %调整图片标题与下文距离
   \centering
      \includegraphics[width=1.00\linewidth]{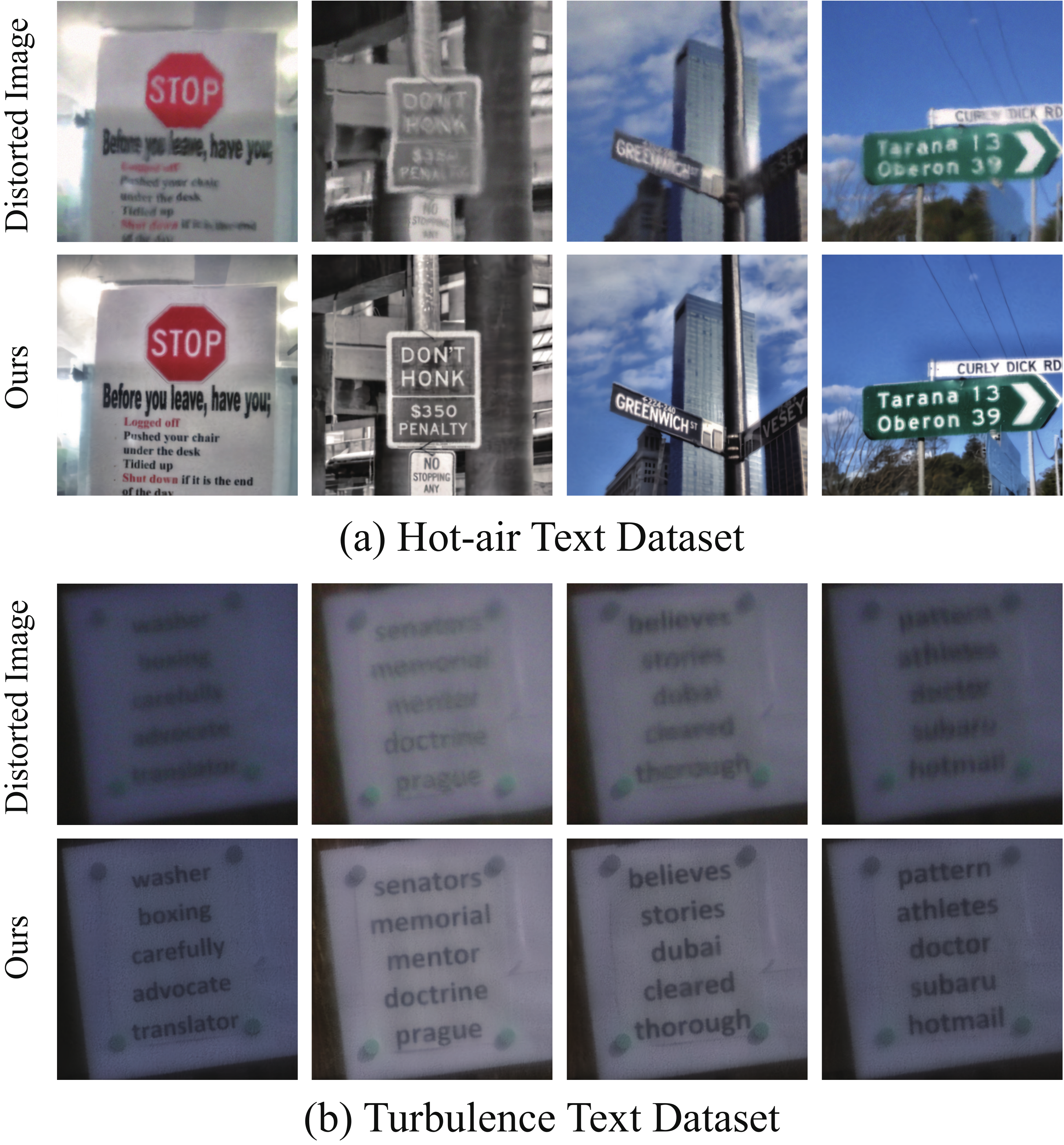}
   \caption{The visual examples of reconstruction results for both types of text dataset. (a) and (b) show the
   results of hot-air text dataset and turbulence text dataset, respectively.}
   \label{dataset}
 \end{figure}
This technical report presents our proposed solutions to CVPR 2023 UG$^{2}$+ Challenge Track 2.1-Text Recognition Through Atmospheric Turbulence. The participant's task is to mitigate the adverse effects caused by the turbulence so that the text recognition system can successfully recognize the text in the restored images.

In this track, two types of text datasets are provided during the final testing phase: hot-air text dataset (Fig. \ref{dataset}(a)) and turbulence text dataset (Fig. \ref{dataset}(b)). The former is generated by simulating physical turbulence on images using a heat chamber, while the latter is obtained from a distance of 300 meters in hot weather \cite{mao2022single}. The hot-air text dataset comprises 400 sequences, and the turbulence text dataset comprises 100 sequences, with each sequence composed of 100 distorted frames. It is required to reconstruct a high-quality image from these distorted frames. The reconstruction result of the final testing phase is based on the average accuracy of three text recognition systems (CRNN \cite{shi2016end}, ASTERN \cite{shi2018aster}, DAN \cite{wang2020decoupled}).

We propose a multi-stage framework to mitigate the distortion caused by the turbulence. Firstly, the sharpest frames are selected using frame selection based on sharpness. Next, each frame in the selected frames is aligned to suppress geometric distortion through optical-flow based registration. Then, we utilized an image fusion method with DT-CWT to mitigate the blur caused by the turbulence. Finally, we apply a learning-based deartifacts method to further improve the image quality. Our framework is capable of handling both types of text datasets, and it achieved 1st place on the final leaderboard.
% Finally, our approach shows mighty competitive ability on restoring on GT-Rain testing sequences in the final phase, which ranks first in terms of SSIM and second in terms of PSNR on the final leaderboard. 

 The technical report is organized as follows: Section \ref{sec2} briefly describes the restoration framework. Experimental results are given in Section \ref{sec3} to show the performance as compared with other methods.
 \begin{figure*}[t]
  \centering
     \includegraphics[width=1.00\linewidth]{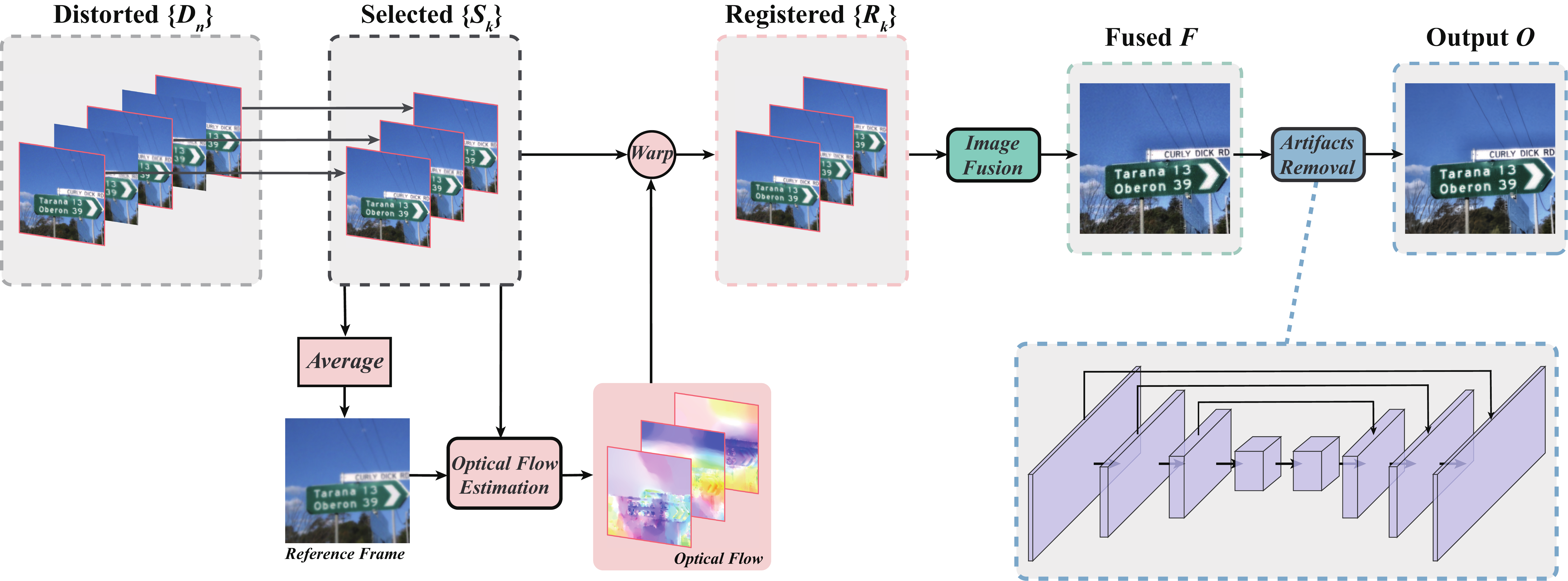}
  \caption{Block diagram for the proposed restoration framework. We firstly utilize
  a frame selection algorithm based on sharpness to select the best set of distorted frames. Next, each
  frame in the selected frames is aligned to suppress geometric
  distortion through optical-flow based registration. Then, a
  region-based image fusion method with DT-CWT is utilized
  to mitigate the blur caused by the turbulence. Finally, a
  learning-based deartifacts method is applied to improve the
  image quality, generating a final out. }
  \label{Framework}
\end{figure*}
\section{Restoration Framework}\label{sec2}
The proposed restoration framework contains four main steps (see the diagram in Fig. \ref{Framework}):

\textbf{A.} Frame Selection;

\textbf{B.} Image Registration;

\textbf{C.} Image Fusion;

\textbf{D.} Artifacts Removal.

\subsection{Frame Selection}\label{sec2.1}
\begin{figure}[tb]
  % %  \vspace{0.2cm}  %调整图片与上文的垂直距离
  % %  \setlength{\abovecaptionskip}{0 cm}   %调整图片标题与图距离
  %  \setlength{\belowcaptionskip}{-0.4 cm}   %调整图片标题与下文距离
     \centering
        \includegraphics[width=1.00\linewidth]{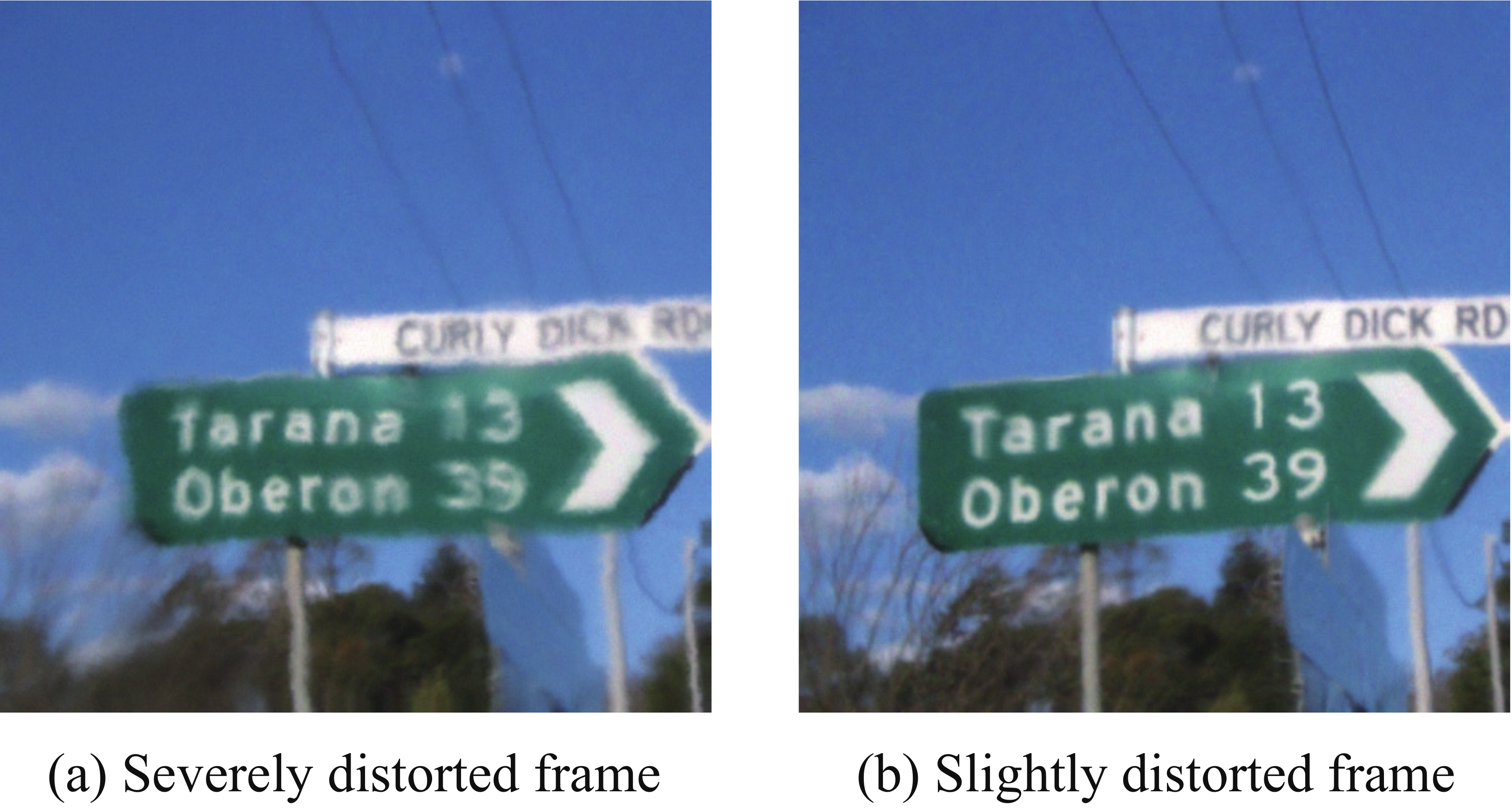}
     \caption{Frames with different visual quality in a sample sequence distorted
     by the atmospheric turbulence. (a) is a severely distorted frame that has a negative contribution in restoring the high-quality image, while (b) is a slightly distorted frame containing more useful information.}
     \label{diff}
\end{figure}

In high-temperature imaging, atmospheric turbulence affects frames in a sequence unequally. The degree of distortion varies from frame to frame due to random fluctuations of the refractive index in the optical transmission path. Consequently, some frames have better image quality than others, with less blurriness and more useful image information.
This point is illustrated in Fig. \ref{diff}, where two sample frames of a signboard scene sequence that is distorted by atmospheric turbulence are shown. Fig. \ref{diff}(a) is a frame severely distorted by the turbulence, while Fig. \ref{diff}(b) is a sharp frame from the same sequence. A comparison of the two frames reveals that Fig. \ref{diff}(a) has a negative contribution to the image restoration.

Given an observed sequence \textbf{\{\textit{D$_n$}\}}, each frame in the sequence distorted by atmospheric turbulence can have varying visual quality. Sharpness is a crucial factor that determines the amount of detail information conveyed by an image. Hence, we utilize a frame selection algorithm based on sharpness to select the sharpest set of distorted frames
\textbf{\{\textit{S$_k$}\}} that can aid in accurately reconstruting a high-quality image. In this step, we compute the sharpness based on the intensity gradients of the image.

\subsection{Image Registration}\label{sec2.2}
In step B, each frame in the selected frames \textbf{\{\textit{S$_k$}\}} is aligned with a reference frame using optical flow, generating a registered sequence \textbf{\{\textit{R$_k$}\}} with less geometric distortion, 
and the reference frame is constructed by averaging the selected sequence \textbf{\{\textit{S$_k$}\}}. The
purpose of this step is to suppress geometric distortion.
\begin{figure}[htbp]
  % \vspace{0cm}  %调整图片与上文的垂直距离
 %  \setlength{\abovecaptionskip}{0 cm}   %调整图片标题与图距离
  % \setlength{\belowcaptionskip}{-0.8 cm}   %调整图片标题与下文距离
    \centering
       \includegraphics[width=1.00\linewidth]{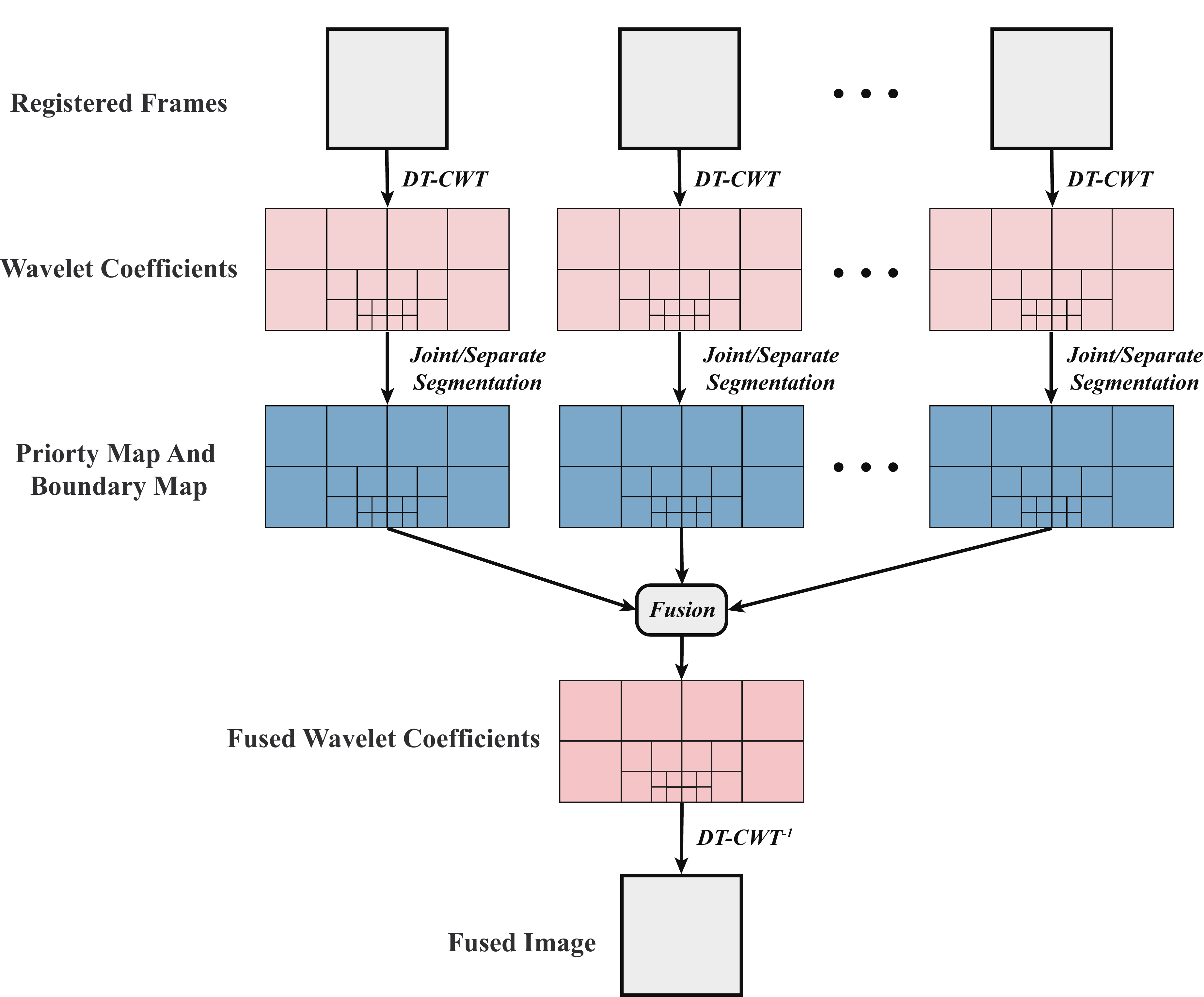}
    \caption{ The Region-Based image fusion scheme using the
    DT-CWT.}
    \label{DTCWT}
  \end{figure}
\subsection{Image Fusion}\label{sec2.3}
\begin{figure*}[htbp]
  % \vspace{-0.6cm}  %调整图片与上文的垂直距离
%  \setlength{\abovecaptionskip}{0 cm}   %调整图片标题与图距离
%  \setlength{\belowcaptionskip}{2 cm}   %调整图片标题与下文距离
 \centering
    \includegraphics[width=1.00\linewidth]{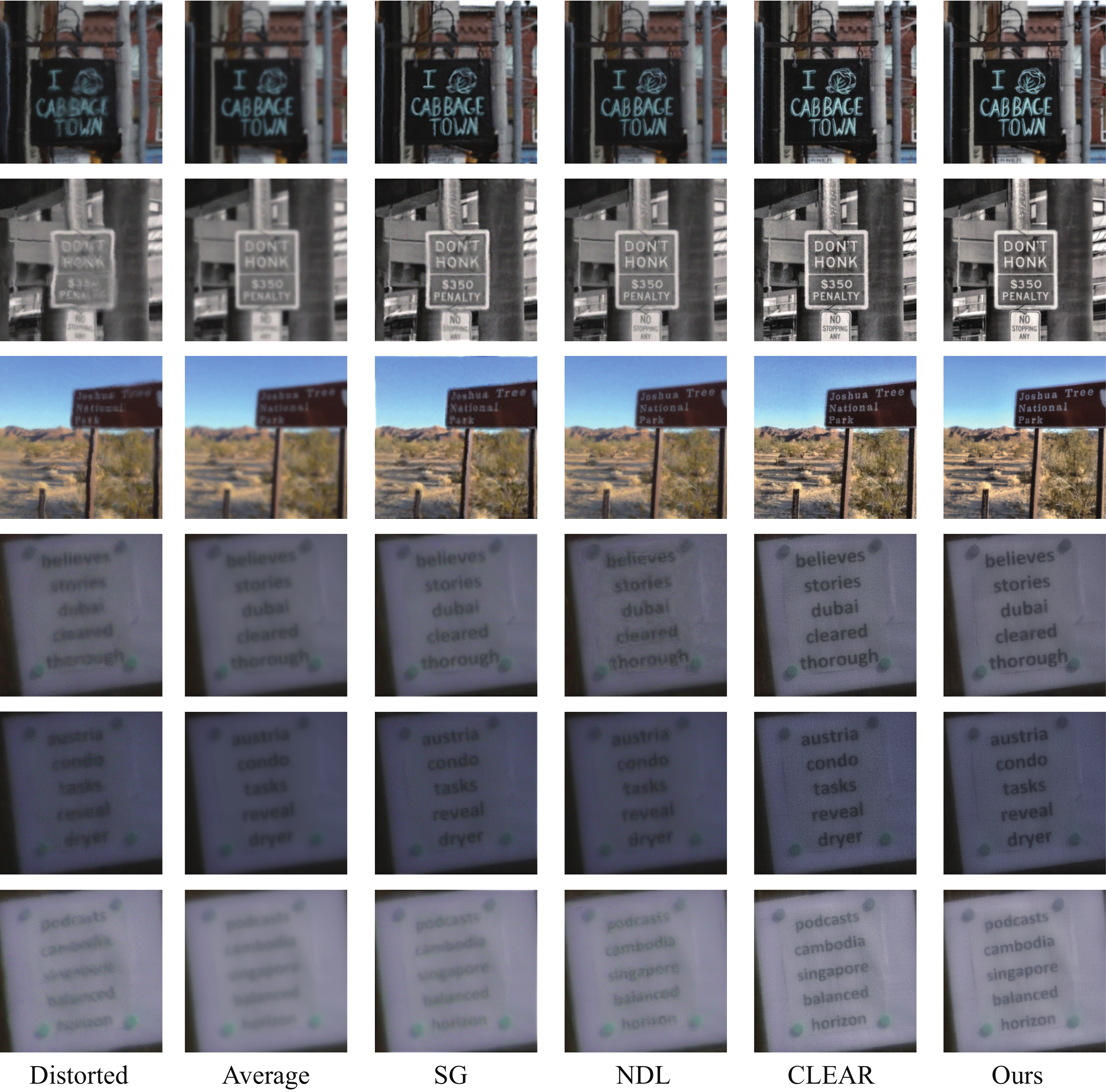}
 \caption{Visualization of reconstruction results on both types of text dataset.}
 \label{comparison}
\end{figure*}
Step C involves in restoring a single image \textbf{\textit{F}} from the registered sequence  \textbf{\{\textit{R$_k$}\}}. We utilize a region-based image fusion method with DT-CWT to mitigate the blur caused by turbulence. The fusion framework is shown in Fig. \ref{DTCWT}. The DT-CWT \cite{lewis2007pixel,lewis2004region,anantrasirichai2013atmospheric} is a popular technique for image fusion due to its shift invariance, orientation selectivity, and multiscale properties, which enables the selection and combination of useful information from multiple source images to generate a new image. However, the fusion process may introduce artifacts that affect the image quality.
 \begin{figure*}[htpb]
  %  \vspace{-0.5cm}  %调整图片与上文的垂直距离
%  \setlength{\abovecaptionskip}{0 cm}   %调整图片标题与图距离
%  \setlength{\belowcaptionskip}{2 cm}   %调整图片标题与下文距离
  \centering
     \includegraphics[width=1.00\linewidth]{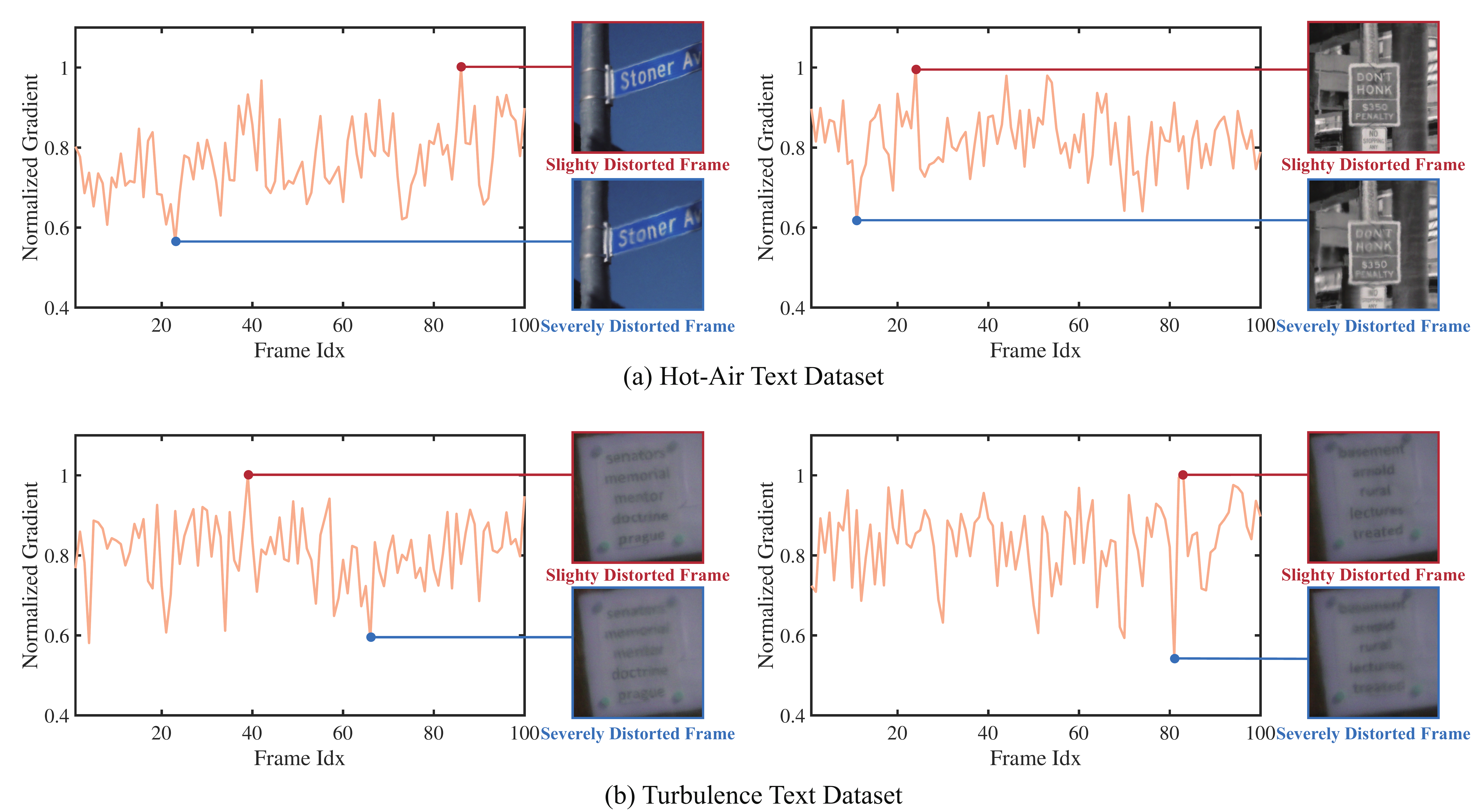}
  \caption{Normalied gradient for both types of text sequences. Both types of text sequences
  exhibit a strong fluctuation in the the temporal normalized
  gradients, which demonstrates that the degree of distortion
  of each frame is different from one another in a high-temperature environment.}
  \label{Analysis}
\end{figure*}

\subsection{Artifacts Removal}\label{sec2.4}
Finally, a learning-based deartifacts method FBCNN \cite{jiang2021towards} is applied to remove the 
remaining artifacts in the Fused image \textbf{\textit{F}}. FBCNN is capable of predicting the quality factor of a JPEG image and embedding it into the decoder to guide image restoration. Besides, the quality factor can be manually adjusted 
for flexible JPEG restoration according to the user's preference. In this task, we 
have set the quality factor to 20.
\section{Experiments}\label{sec3}

\subsection{Datasets and Experimental Setting}\label{sec3.1}
\noindent\textbf{Datasets.} We conduct experiments on both hot-air text dataset and turbulence text dataset to evaluate the proposed framework.

\begin{itemize}
  \item \textbf{Hot-air text dataset.} Hot-air text dataset is generated
  by simulating physical turbulence on images using a heat chamber, which contains 400 sequences in this track.
  \item \textbf{Turbulence text dataset.} Turbulence text dataset is obtained from a distance of 300
  meters in hot weather, which contains 100 sequences in this track.

\end{itemize}

\noindent\textbf{Experimental Setting.} We compare the proposed framework with several existing atmospheric turbulence mitigation methods, including CLEAR \cite{anantrasirichai2013atmospheric}, SG \cite{lou2013video}, NDL \cite{zhu2012removing}.

\subsection{Experiments On Text Datasets}\label{sec3.2}
We evaluate the performance of the proposed framework and some existing methods on both types of text dataset. The visual results of different methods are shown in Fig. \ref{comparison}.
As for visual comparison in the results, NDL \cite{zhu2012removing} and SG \cite{lou2013video} struggle to handle the turbulence text dataset and fail to correct geometric distortion. Albeit CLEAR \cite{anantrasirichai2013atmospheric} can effectively mitigate the blur and geometric distortion caused by turbulence, it introduces the severe artifacts into the results. 
Compared with these methods, the proposed framework outperforms these methods by simultaneously mitigating the blur and geometric distortion without destroying image details, demonstrating superior performance.

\subsection{Discussion}\label{sec3.3}
\noindent\textbf{Analysis of the image quality of different frames.} In order to investigate whether atmospheric turbulence affects each frame differently in a high-temperature environment, we analyzed the image quality using the normalized gradient as a sharpness metric for each frame in the distorted sequences.
As shown in Fig. \ref{Analysis}, we can observe a considerable fluctuation in the temporal normalized gradients for both types of text sequences, indicating that the degree of distortion for each frame varies. Therefore, some of the frames have better image quality than others,
with less blurriness and more useful image information.
\section{Conclusion}
In our submission to the track 2.1 in UG$^{2}$+ Challenge in CVPR 2023, we propose an efficient multi-stage framework to restore a high-quality image from distorted frames. Firstly, A frame selection algorithm based on sharpness is first utilized to select the best set of distorted frames.
Next, each frame in the selected frames is aligned to suppress geometric distortion through optical-flow based registration. Then, a region-based image fusion method with DT-CWT is utilized to mitigate the blur caused by the turbulence. Finally, a learning-based deartifacts method is applied to improve the image quality, generating a final out. Our framework can handle both hot-air text dataset and turbulence text dataset provided in the final testing phase, 
and ranked 1st in text recognition accuracy. In the future, we will
explore more efficient methods to improve this task.

{\small
\bibliographystyle{plain}
\bibliography{refpaper.bib}
}

\end{document}